# Improved Churn Causal Analysis through Restrained High-dimensional Feature Space Effects in Financial Institutions

David Hason Rudd [1, *], Huan Huo [1] and Guandong Xu [1,2]

## Abstract

Customer churn describes terminating a relationship with a business or reducing customer engagement over a specific period. Customer acquisition cost can be five to six times that of customer retention, hence investing in customers with churn risk is wise. Causal analysis of the churn model can predict whether a customer will churn in the foreseeable future and identify effects and possible causes for churn. In general, this study presents a conceptual framework to discover the confounding features that correlate with independent variables and are causally related to those dependent variables that impact churn. We combine different algorithms including the SMOTE, ensemble ANN, and Bayesian networks to address churn prediction problems on a massive and high-dimensional finance data that is usually generated in financial institutions due to employing interval-based features used in Customer Relationship Management (CRM) systems. The effects of the curse and blessing of dimensionality assessed by utilising the Recursive Feature Elimination (RFE) method to overcome the high dimension feature space problem. Moreover, a causal discovery performed to find possible interpretation methods to describe cause probabilities that lead to customer churn. Evaluation metrics on validation data confirm the Random Forest and our ensemble ANN model, with %86 accuracy, outperformed other approaches. Causal analysis results confirm that some independent causal variables representing the level of super guarantee contribution, account growth, and account balance amount were identified as confounding variables that cause customer churn with a high degree of belief. This article provides a real-world customer churn analysis from current status inference to future directions in local superannuation funds.

# 1. Introduction

Businesses rely heavily upon retaining satisfied customers in the global marketplace, which account for a large portion of their revenue. As the market becomes increasingly saturated, businesses have learned to emphasise maintaining existing clients. Although obtaining new clients was critical for initial business success, retention policies should have equal weight. Many previous studies have identified retention rate's substantial effect on the market, but clients can always churn away from a business, resulting in potential losses for the organisation. However, customers usually offer some warning before being churned. Hence, churn prediction systems primarily focus on customer behaviour to identify specific customers who are likely to churn out and indicate reasons for the churn. Such factors would aid marketing to develop effective retention strategies, increasing overall customer lifetime value, and assisting in growing the company's market value. For example, applying incentive programs for customers who had fewer transactions/interactions over time. In general, churn rate is calculated by dividing the number of lost customers by initial customers [1].

Voluntary churn is when a consumer wants to leave the company on their own. This could be due to dissatisfaction with the product, or perhaps they feel they are not receiving that value they expected. In contrast, involuntary churn is where a customer leaves the company for unavoidable reason(s), such as payment problems due to expired account details, network issues, or inadequate cash, and incidental churn occurs due to location or financial situation changes. Deliberate churn arises due to customer desire to change innovation and price, where the most common reasons are poor service quality, non-competitive pricing, and missing customer expectations. Companies can utilise churn analysis to determine the individual user risk levels and develop appropriate, focused retention initiatives.

*Corresponding Author: David Hason Rudd (david.hasonrudd@student.uts.edu.au)
[1] The University of Technology Sydney, 15 Broadway, Ultimo, Australia
[2] Data Science Institute, 15 Broadway, Ultimo, Australia

The proposed framework applies churn analysis for a local superannuation fund. Data preparation and analysis verified that most customer accounts were active for less than one year, for several possible reasons, including some customers being involuntary churners who were subsequently enrolled in a different superannuation fund(s) operated by their employers. In contrast, voluntary churn occurs when a customer decides to cease the account for personal reasons, including perceived quality, technology, and price. The findings from this study make several major contributions to the current literature, such as:

- This study is the only empirical investigation into the impact of a causal effect on churn in superannuation funds.
- The present study has enhanced our understanding of exploiting correlation discovery tools within the causal inference of churn.
- This work has combined different approaches such as RFE, SMOTE sampling techniques, ensemble ANN, and Bayesian networks within a practical framework to explore hidden parameters that cause a churn to occur.

First, we analysed 12-month time window data to predict churners for the next six months. Since mining data from the 12-month historical data allowed us to extract the latent factors that cause churn. Then, we scaled up the model by exploiting Bayesian networks to describe how deliberate churn occurs due to multiple causes and demonstrated that hypothesis tests on common features greatly influence prediction results. Finally, we proposed a specific causality analysis method that can be applied to other similar datasets employed in most superannuation funds.

It should be noted that this study only focuses on the performance of a DFF NNs to address churn prediction on a massive and high-dimensional sparse dataset that is usually created in financial institutions. These very high dimensions data are primarily generated in subscribed-base businesses, mostly in superannuation fund(s) due to employing interval-based features used in Customer Relationship Management (CRM) systems. The Recursive Feature Elimination method is employed to overcome the high dimension feature space problem, and then compared the results with proposed ensemble ANN and other classifiers.

The remainder part of the article proceeds as follows: Section 2 presents a comprehensive literature review of recent relevant works on customer churn problem, and section 3 gives preliminary knowledge on the application of the deep learning method in churn prediction problems. Section 4 discusses specific research and analysis methods employed to predict underlying reasons for churn including problem definition and churn propensity modelling workflow. Section 5 setup experiment analyses on gathered data to address research questions: a causal analysis of churn through the financial data collected by superannuation funds and section 6 presents results on prediction and causality analysis outcomes and discusses implications from the findings for real-world applications and limitations. Finally, section 7 summarises and concludes the article.

# 2. Related Work

## 2.1 Customer Churn Prediction

Different churn prediction techniques have been evaluated to identify optimal approaches [7]. Most previous studies focused on determining churn variables for a particular dataset rather than customer churn causation analysis.

The state-of-the-art churn prediction framework is designed, which is predicated upon "deep neural models, time-to-next-event models, and Big Data processing" using large parallel computing with GPU units [22]. Developing the predictive model of customer behaviour in order to plan for and handle such situations could be quite beneficial. Staff churn and staff loss would be similar to customer churn, however the effect of losing a significant customer for an organisation will almost certainly be even more stressful (since organisations have no physical sense of losing their staff), whereas the implications of seeking well employees rather than missed employees, and also the expense of in-service coaching which should be provided for the new hires, might be significant [30]. Using the Partial Least Squares (PLS)-based technique upon highly linked sets of data across variables, create accurate and compact predictive models for churn prediction [27]. The use of hybrid learning algorithms towards churn prediction in mobile networks allows for predictive modelling of buyer behaviour [28]. A Locally Linear Model Tree (LOLIMOT) method combines the benefits of neural networks, fuzzy modelling, and tree models [29]. Moreover, the RemsProp training approach outperformed "conventional Stochastic Gradient Descent (SGD), Adam, AdaGrad, Adadelta, and AdaMax" algorithms in terms of accuracy for deep learning-based churn prediction [1].

The effectiveness of a typical random forests strategy for predicting customer churn was studied, and the integration of sampling approaches and price learning into the method to outperform many proposed algorithms used a real bank dataset [25]. Based on Orange Belgium customer information, the real-life churn prediction case gets constructed. The first section of the paper involves the creation of such a precise prediction model. An Easy Ensemble technique uses a random forest classifier to address a considerable class imbalance between two categories [26].

For non-subscription-oriented business sets, a method for predicting customer churn was suggested. The set of general features can be derived from practically all non-subscription-oriented firms' revenue and transaction information and used to forecast customer churn. For prediction, "a neural network-based Multilayer perceptron" is often used [11].

CHAMP (Churn analysis, modelling, and prediction) is an integrated system for forecasting consumers cancelling their cellular phone service [3]. Alyuda NeuroIntelligence employs Neural Networks (NNs) for data mining to forecast customer churn at banks [4]. Integrating textual data using Customer Churn Prediction (CCP) algorithms adds value [5], and combining different classifiers, e.g., gradient boost, oversampling, and contrast sequential pattern mining on single-year observation windows have been demonstrated to be a practical strategy to deal with highly skewed data collected from superannuation funds [6].

Hidden churn is a common problem for superannuation funds, where customer accounts become dormant once mandatory employer payments cease. Various remedies for insufficient consumer interaction have been proposed, addressing unbalanced and fully leveraged data problems, and multiple classifiers have been developed from sampled datasets [6]. In addition, deep learning techniques can handle large datasets compared with standard Machine Learning (ML) approaches and combining deep learning and Convolutional Neural Networks (CNNs) has successfully forecasted churn [8].

## 2.2 Studies on Causal Inference of Churn

Recent causal inference developments highlight fundamental alterations required to move from standard statistical analysis toward multivariate causal analysis [3]. New PC-Stable approaches effectively learn causal structures using DFS data, allowing temporal causal modelling for enormous time information datasets [9]. Directed Acyclic Graph (DAG) has been proposed to represent causal relationships in Bayesian networks [10]. A likelihood of churn anticipates, as are the driving factors on banking data. Subsequently, Shah et al. [11] trained a model to produce suitable weights for features that predict whether a customer would churn. They contrasted churn definitions commonly used in business administration, marketing, IT, telecommunications, newspapers, insurance, and psychology. Many studies have described churn loss, feature engineering, and prediction models using this approach [2]. The text content from daily Twitter posts was analysed using a Convolution Neural Network to determine if the sentiment was good, bad, or indifferent. Granger causality analysis has been used to cross-validate that generated mood data set [23].

According to our literature review, no research has been linked with causal analysis on client attrition in superannuation fund(s). Moreover, the above approaches have been variously adapted for current churn prediction methods. However, most researchers evaluate models based on telco, media, retail, medical, and insurance data. In contrast, the proposed approach was tested using a massive financial dataset with high-dimensional sparse data associated with a local superannuation fund. Motivated by the studies mentioned above, we proposed a new approach that combines different algorithms including RFE, SMOTE, ANNs deep learning and Bayesian Networks to improve churn propensity modelling.

# 3. Preliminary Knowledge

Churn prediction models will provide earlier churner identification and hence assist customer intervention program development. This is basically a classification problem, i.e., to categorise each customer as a potential churner or non-churner. Several machine learning techniques for churn prediction have produced verifiable results from interpretable models, such as boosting technique, a non-parametric method, and logistic regression. However, this approach is only valid for predictability under circumstances where the customer database size is minimal, varying sample size, and fails for larger datasets [7]. Therefore, we propose a Deep Learning (DL) algorithm to deal with massive financial data volumes, since deep learning (feature transformation) differentially weights features using historical data. Deep learning algorithms follow human brain architecture using Artificial Neural Networks (ANNs) [12]. Feed-forward NNs use the output from one layer as input for the next layer, with no loops between layers [13]. Therefore, the Deep Learning Feed-Forward (DFF) or ANNs algorithm is employed to deal with high-dimensional sparse data. General advantages for ANNs DL can be summarised as follows:

1. superior accuracy from DL;
2. includes more variables than classical ML;
3. DL algorithms can extract patterns while avoiding blind spots from extensive customer demographics, behavioural variables, and billions of customer engagement logs;
4. reduces time-consuming feature engineering and manual financial data analysis.

One significant weakness of deep learning is that it remains a black-box model, i.e., DL does not express uncovered patterns in the underlying data in easily understandable ways. However, to address the complexity of the ANNs model, a causal inference model is proposed to identify churn effects of treatments and describe cause probabilities that lead to churn.

# 4. Methodology

## 4.1 Data Mining and Classification Problem

Data mining can identify useful knowledge in terms of pattern extraction from different sources, and various feature engineering tools can extract hidden patterns from massive datasets [12,14]. Figure 1 demonstrates that this study included 12 datasets from a superannuation company for real-world experiments to verify the proposed model's effectiveness.

This study addresses the classification problem of minority class and overfitting due to the massive dataset. The classification problem is limited to two classes. A vector includes input data as each member has $n$-components or features. The pattern of each member's data is $P$ with $n$-dimensional feature space in class 1 (minority class) or class 0 (majority class). So, a training set of vectors $\{x1, x2, x3, ... xk, ... xn\}$ with class label of $\{y1, y2, y3, ... yk, ... yn\}$, so that $yk \in \{0,1\}$ is defined to recognise the vector of $n$-components or patterns a decision boundary is defined in a discriminant function named $D(x)$ into decision boundary of $D(x) > 0$ and $D(x) < 0$ to assign each sample to a churn (1) or non-churn (0) class as demonstrated in equation (1):

$$D(x) > 0 \, \exists \, x \in class(0) \Leftrightarrow acc\_close\_t_W > acc\_close\_t_W - 1 + t_w$$

$$D(x) < 0 \, \exists \, x \in class(1) \Leftrightarrow acc\_close\_t_W \leq acc\_close\_tw - 1 + t_w$$

$$D(x) = 0 \quad or \quad acc\_close\_t_w = t_w \quad \text{decision boundary} \tag{1}$$

where $x$ is an input pattern and $t_w$ denotes a 6-month time window, $acc\_close\_t_w$ represents current time window, and $acc\_close\_t_w - 1$ demonstrates previous 6-month time window. Thus, we built a simple linear discriminant function by calculating the sum of the training patterns and bias as shown in (2):

$$D(x) = w.x + b \tag{2}$$

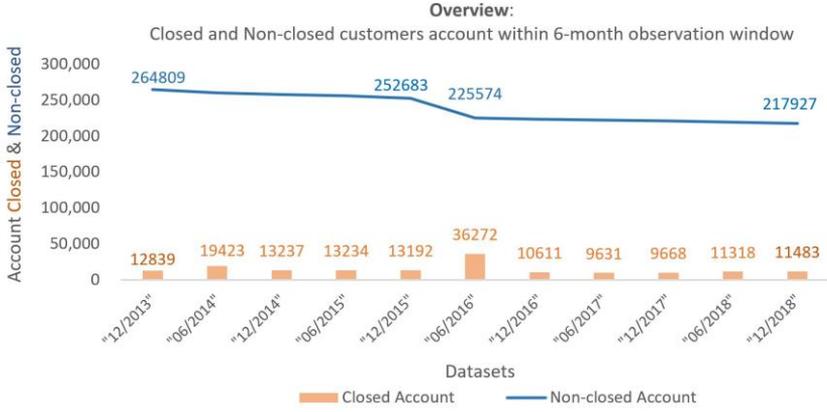

Fig. 1: Closed and non-closed customer accounts in periods of 6-month.

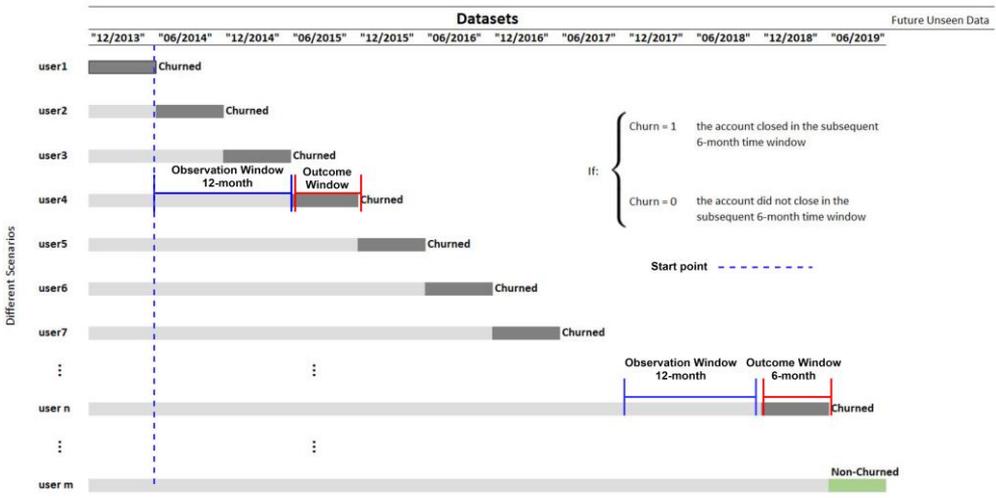

Fig. 2: Extracting feature values by sliding observation window on datasets. Observation and Outcome time-windows set to 12-months and 6-months respectively.

where $w$ represents weight of pattern and $b$ symbolises bias. Since a singular criterion for decision boundary is defined, a linear discrimination function can separate the classes without error.

As mentioned in Equation 1, we define a customer as a churner if they closed their account during the subsequent 6-month time window. Therefore, we use a binary outcome for each customer [0 or 1], where 1 means the account closed and 0 that it was not close in the subsequent 6-month time window.

In churn analysis, the data mining method should fit the problem structure based on the definition of active and inactive account holders in financial institutions. Two main inclusion criteria are defined to satisfy dimensionality reduction by eliminating massive noisy data. First, customers with more than six months of account tenure are only retained. Second, account balances below \$1500 are considered low engagement members and removed. Figure 2 demonstrates features in the observation window to predict which user will be churned or non-churned in the next 6-month outcome window.

## 4.2 High Dimension Feature Space

The dataset included more features than observations for each member results risk of overfitting in the model and occurring in high dimensional feature space, which is common in finance data. On the other hand, many features have low variance and correlation with the target variable. The model performance of causal analysis and causal effect of treatment outcomes depends on observing all available causal variables, possible covariates, and cofounders against dropping low importance feature increase bias in causal model. Therefore, we should trade-off between the effect of the curse of dimensionality [32] and the blessing of dimensionality [31]. To achieve this, different dimensionality reduction methods are investigated including feature dropping, wrapper methods and feature importance with random forest. Finally, the Recursive Feature Elimination RFE [33] obtained robust results to overcome high dimensional data while retaining the possible influential cofounders and causal variables. The RFE is a popular feature-ranking algorithm to remove low-weight features since it gives us control to set threshold and fixed number of top-ranked features.

RFE removes the feature with the smallest ranking criterion using the cost function DJ(i) is demonstrated in (3). The feature weight assesses changing cost function output in weight $D\omega_i = \omega_i$ by eliminating a given feature $i$ in an iterative procedure called RFE [34].

$$DJ(i) = (1/2)\frac{\partial^2 J}{\partial w_i^2}(Dw_i)^2 \qquad (3)$$

Where DJ(i) in (3) ranks weight criterion to expand J in Taylor series to second order, and J in (4) denotes cost function in classifier [33].

$$J = \sum_{x \in X} |w \cdot x - y|^2 \qquad (4)$$

The features remove in into subset $Fm$ each iteration from the lowest to the highest ranked features. Therefore, we proposed a method to combine SMOTE and RFE algorithm to overcome minority class and high dimension feature space problem so that features with low variance value and predictive power are removed from the data without significant impact on the causal inference model.

## 4.3 Churn Propensity Modelling

We compared the proposed DFF NNs algorithm approach with seven current best-practise classifiers to evaluate the proposed framework's effectiveness on the datasets with 12-month observation window and 6-month test window, as discussed in problem definition section. The workflow of the proposed model demonstrates in Figure 3. After feature selection, the proposed algorithms were employed for training (80%), and test (20%) sets to derive 193 features from 124363 total examples. Binary classification generally produces severely skewed data distributions. Hence, we employed the Synthetic Minority Oversampling Technique (SMOTE) [17] in pre-processing to synthesise new examples for the minority class to ensure equal sample count (14031) for each class (1 or 0) in the training set. Experimental results confirmed improved performance for the proposed model using the sampling method.

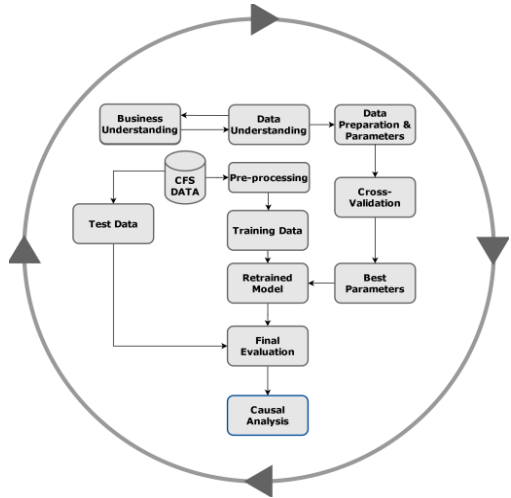

Fig. 3: Churn propensity modelling

Furthermore, a majority voting ensemble from the Scikit-learn library is employed [18] to combine predictions from multiple models, which can be helpful to improve model performance for classification tasks. Ensemble hard and soft voting were both applied for the comparison supervised models. Hard voting counts each individual classifier votes and the majority wins; whereas soft voting weights each prediction by classifier importance, and the target label with the greatest sum of weighted probabilities wins [18]. In addition, a specific ANNs architecture is built to address the research problem using ensemble ANNs networks. Table 1 shows the ensemble ANN's architecture optimised based on the hyperparameter tuning analysis. Finally, in the last step of the churn propensity model, we considered features that have the most predictive power on classifier outcome as possible causal effect by analysing model output with two popular plots named SHapley Additive Explanation (SHAP) and Partial Dependence Plot (PDP). Therefore, the initial assumption is generated by observing how a feature affects predictions and what is the relationships between influential features and prediction accuracy. Then, our assumptions are represented in a DAG based on the previous knowledge obtained from correlation analysis and evaluating influential predictors. Therefore, we take advantage of both Machine Learning (ML) and Bayesian systems in this study. In ML, we predicted the target or independent variable by observing dependent variables, selected the most influential features to enhance prediction accuracy, and used them in our proposed causal inference model. Therein after, a novel approach for causal analysis with robust outcome is applied in this study. We built a dependency structure between dependent features, i.e., a causal graph, based on statistical relationships between independent features. Different methods to identify causal traces from large datasets enabled predicting more flexible causal relations [19].

Table 1. Network Architecture of Ensemble ANNs

| Network type | Deep ANN-1 | Deep ANN-2 |
|---|---|---|
| Hidden layers | 4 | 4 |
| Dense activation 1,2,3 and 4 | tanh, and 3 x relu | tanh, and 3 x relu |
| Dropout 1, 2,3 and 4 | 0.2, 0.2, 0.2, and 0.2 | 0.4, 0.4, 0.4, and 0.4 |
| Output activation function | Sigmoid | Sigmoid |
| Learning rate | 0.000474718 | 0.000012 |
| Epochs | 100 | 100 |
| Batch size | 512 | 512 |
| Optimisation algorithm | ADAM | ADAM |

# 5. Experiment Setup

## 5.1 Datasets

Experiments were conducted on 12 datasets for members holding accounts provided by a local finance company. All 12 datasets included customer accounts, demographics, customer engagement, and financial data, with approximately 250,000 examples with 88 features (71 numerical and 17 nominal) in each dataset. Figure 4 visualises customer and account tenure distribution data that are almost identical in all datasets. Therefore, it can be inferred that approximately 90% of customers had 5–12 months account tenure, with an average tenure of $\approx$ 7.6 months. Thus, most customers closed their accounts in less than one year. Therefore, we adopted the rational approach to only retain customer accounts that were open for more than six months, i.e., we eliminated all recently opened accounts.

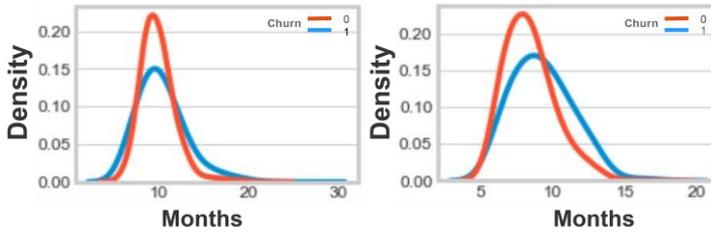

Fig. 4: Distribution of customer and account tenure

## 5.2 Evaluation Metrics

Measuring model performance is essential for ML, and several techniques evaluate model effectiveness for regression and classification tasks. We used Area Under the Curve (AUC) and recall visualising overall performance of a classifier. AUC represents how capable the model is to discriminate classes [15], with larger AUC indicating better distinguishability between churned and non-churned customers, and AUC $\approx$ 1 represents good separability. AUC is defined as [6] and shown below equation,

$$AUC = \frac{1}{mn} \sum_{i=1}^{n} \sum_{j=1}^{m} 1 . pi > pj \qquad (5)$$

where $i$ denotes the whole data points from 1 to m when the churn label 1 in churned class $pi$ and $j$ denotes all data points that represent from 1 to n when the churn label 0 in non-churned class. Both $pi$ and $pj$ are probabilities outcomes related to each class here; 1 is an indicator when the condition $pi > pj$ is true.

## 5.3 Correlation Analysis

Relationships between variables can be measured as their correlation defined as [16]:

$$Cor(X,Y) = \frac{\sum_{i=1}^{n}(x_i - \bar{x})(y_i - \bar{y})}{\sqrt{\sum_{i=1}^{n}(x_i - \bar{x})(y_i - \bar{y})}} \qquad (6)$$

The correlation analysis identifies potentially meaningful connections between variables and is applied to select highly correlated relationships for subsequent causal discovery.

## 5.4 Causality Analysis Experiment Setup

We employed Bayesian causal graphs to encode assumptions and determine dependency levels between features, using the DoWhy python package [20]. DoWhy performs causal discovery on all potential ways to identify a desired effect based on the Bayesian causal graph model, exploiting graph-based criteria to find possible interpretation methods [20].

# 6. Results and Discussion

## 6.1 Prediction Results

Figure 5 shows experiment outcomes conducted on the most updated observation and outcome windows. The RF outperformed the other algorithms with AUC = 80%, whereas the proposed ensemble ANNs and Random Forest (RF) have almost the same performance, with improved AUC by 7.5% compared with logistic regression. Thus, ensemble ANNs outcomes were comparable with current best-practise classifiers and achieved maximum prediction accuracy on test data. These findings provide a solid evidence base for exploiting DL reduced time-consuming feature engineering requiring expert knowledge on these specific financial datasets. Furthermore, as shown in Table 2, the low Cohen Kappa Score of 0.86 for the proposed model well proves that there is no big difference between the null error and test accuracy results. Moreover, reliability of model output was measured by the Matthews Correlation Coefficient (MCC) at an acceptable level of 0.45 for the proposed algorithm.

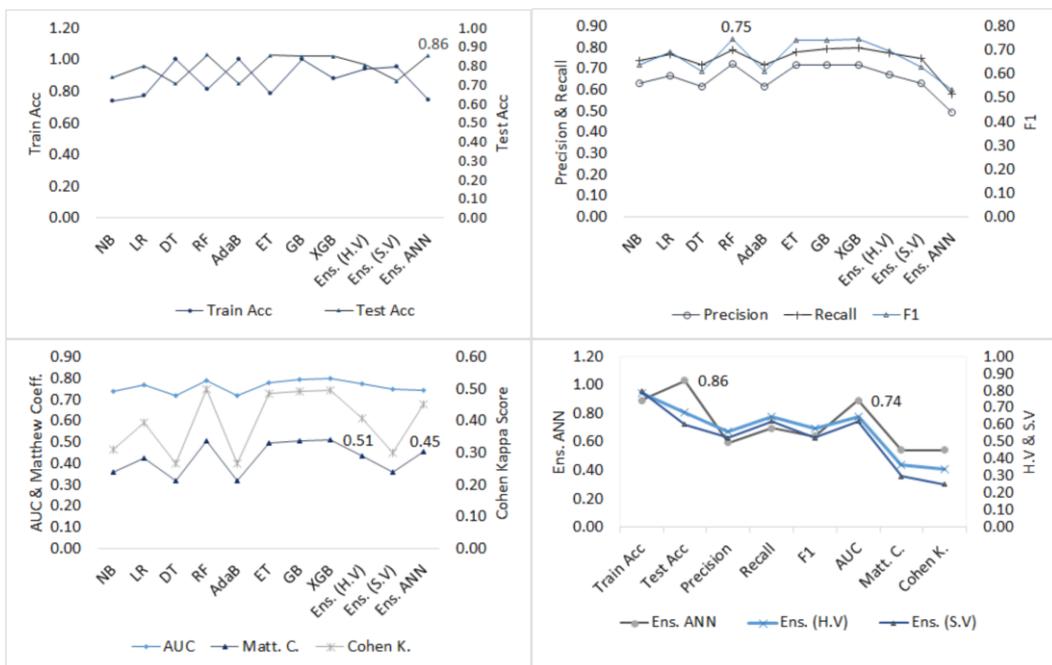

Fig. 5: Several evaluation metrics are applied to the proposed ensemble ANNs model and best-practise classifiers. The ensemble ANNs obtained the highest test accuracy.

Descriptive approaches in statistical analysis define feature weights, reflecting their contribution to pushing model output from its base (average output on the training dataset) to more meaningful outputs. Figure 6 demonstrates SHAP, representing feature impacts on model output, where red features increase and blue reduce prediction outcomes relative to the base value, e.g., feature acc_balance_change_amount reduced and sg_recency increased prediction outcomes. Furthermore, the PDP plot shown in Figure 7 is employed to assess the impact of the feature assumed as a causal effect on the prediction outcome and its relationship with a target variable.

Table 2: The result of test accuracy, Cohen Kappa Score, and Matthew Correlation Coefficient are presented in ten different classifiers.

| Model | Test ACC | Cohen Kappa | Matthew Coeff. |
|---|---|---|---|
| Naive Bayes | 0.74 | 0.36 | 0.31 |
| Logistic Regression | 0.80 | 0.43 | 0.40 |
| Decision Tree | 0.71 | 0.32 | 0.26 |
| Random Forest | 0.86 | 0.51 | 0.50 |
| AdaBoost | 0.71 | 0.32 | 0.27 |
| ExtraTrees | 0.86 | 0.49 | 0.49 |
| GradientBoosting | 0.85 | 0.51 | 0.49 |
| XGboost | 0.85 | 0.51 | 0.50 |
| Stack Ensemble (H.V) | 0.81 | 0.44 | 0.41 |
| Stack Ensemble (S.V) | 0.72 | 0.36 | 0.30 |
| Ensemble ANNs (Proposed) | **0.86** | 0.45 | 0.45 |

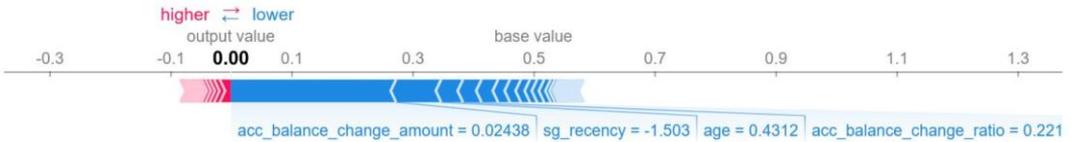

Fig. 6: The above SHAP graph illustrates the feature's impact on prediction outcomes. The promotion_pref and acc_tenure are the most influential features pushing the prediction higher, unlike feature acc_balance_change_amount and sg_recency, which pulls the model's prediction lower.

The impact of the most potent predictors on the model performance is shown in Figure 7. The result of PDP and SHAP plots prove that acc_ballance_change_ratio, login_recency, acc_tenure, cust_tenure, and account_growth_change can be a robust causal estimands in the DoWhy under our assumptions.

## 6.2 Causality Analysis Results

Figure 8 demonstrates empirical results and related causal graphs based on assumptions that would affect churn as follows:

- High-limit account balance might affect the churn as users with a lower limit (account balance) might not be loyal to the superannuation funds compared to customers with a high account balance.
- The account balance change amount could affect the customer tenure. The customer tenure is often based on account balance, after all. The account balance amount itself might affect the churn. An account balance should not be below $1500 in most superannuation funds, and a low account balance generally represents an inactive customer who may be willing to churn.
- A cascade relation between account balance and gender shows that gender would affect account balance and then indirectly affect churn.
- Account growth and high balance change could directly affect churn. No variation in account growth indicates stopped business for most accounts with consequential increased churn propensity.

We identified treatment causal effects on churn outcome based on the initial assumptions by holding other potential effects constant while changing the target treatment. For instance, Linear regression estimation indicates that estimated effect = −0.033853 corresponds to churn probability reducing by ≈ 3% when the customer has lower account growth rate.

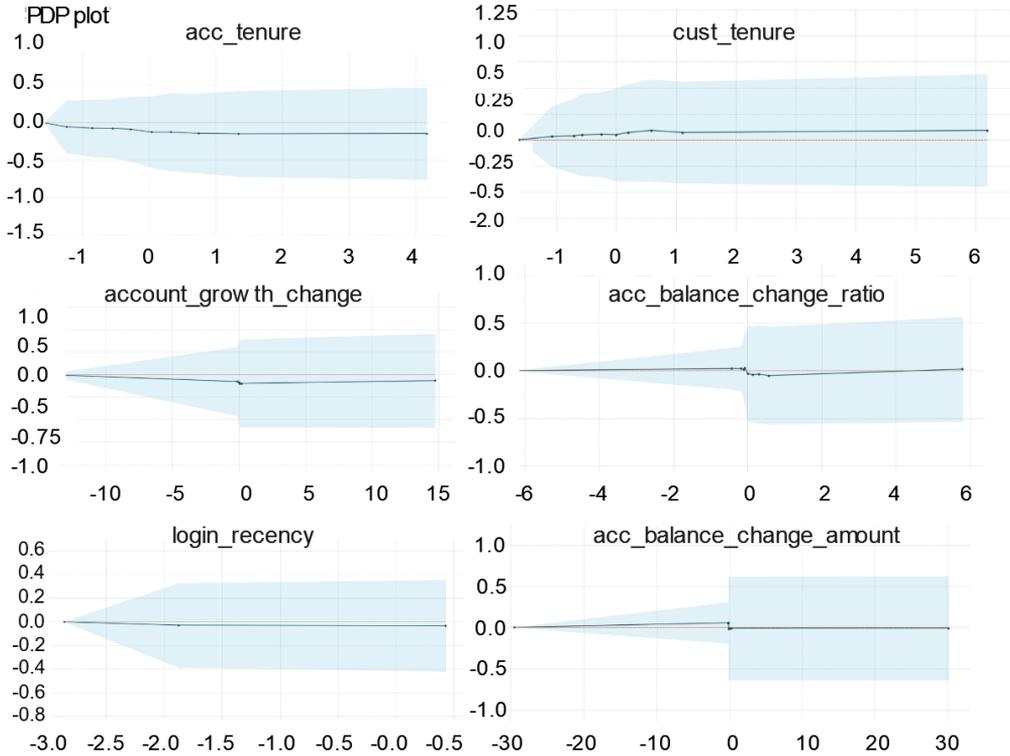

Fig. 7: The six features are identified as the highest predictive power variables on the model's predicted outcome and visualised in the above PDP plot.

To test our assumption so that if the assumption is correct, the new estimation effect should not significantly change. Therefore, we applied Data Subset Refuter (DSR) [20] to refute the above estimates by rerunning them on a random subset of the original dataset. Outcome from the refuting method = −0.033920, almost identical to the estimation result. Thus, we can confirm that the assumption was correct that high account tenure was a causal feature for churn outcome.

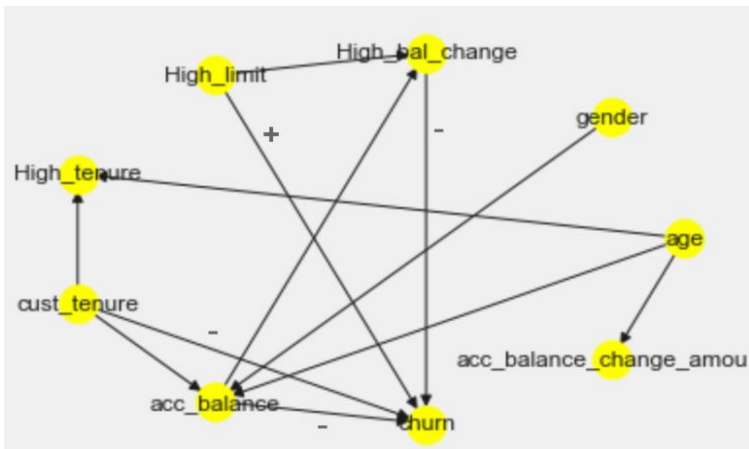

Fig. 8: Causal graph exhibiting assumptions encoded in the causal model

The treatment's causal effect on the outcome is based on the change in the value of the treatment variable. How strong the effect is a matter of statistical estimation. There are many methods for the statistical estimation of the causal effect. In this study, we used the "Propensity Score-Based Inverse Weighting" method [21] and concluded the estimations and churn probability results in Table 3. For instance, the mean estimation for variable sg recency is $\sim 0.15$, which is equivalent to saying that the probability of churn is increased by $\sim 15\%$ when the customer has higher days since the last day of super guarantee SG contribution. The mean estimate of $\sim 0.03$ for the account growth variable can be concluded that churn probability increases by $\sim 3\%$ when the customer has a negative account growth rate. Although the causal analysis result has successfully demonstrated that our assumptions to identify confounding factors are correct with a high degree of belief, it has certain limitations in terms of analysing the identified confounding effects with other popular causal inference methods like the counterfactual analysis.

Table 3: The result of causality analysis illustrates that the assumptions that have a causal effect on customer churn are valid.

| Causal Variable | Estimate Effect | Data Subset Refuter | Probability of churn |
| --- | --- | --- | --- |
| high_bal_change_amount | -0.123401 | -0.122474 | decreased by ~12% |
| high_acc_balance | -0.091698 | -0.091612 | decreased by ~9% |
| low_account_growth | -0.033853 | -0.033920 | increased by ~3% |
| annualrpt_pref | 0.144440 | 0.144457 | increased by ~14% |
| stmt_pref | -0.142732 | -0.142614 | decreased by ~14% |
| high_cust_tenure | -0.027969 | -0.081893 | decreased by ~3% |
| high_sg_recency | 0.156396 | 0.156396 | increased by ~15% |
| promotional_pref | -0.086401 | -0.088061 | decreased by ~8% |

# 6. Conclusion

Losing customers is inevitable for most businesses, but churn can be managed at acceptable levels by investing in customers with the risk of churn. A novel churn propensity model was built and integrated with the causal Bayesian networks. Unbalanced churned and non-churned classes were levelled in pre-processing with SMOTE sampling methods, and then accuracy was compared between the proposed ensemble ANNs and ten best-practise classifiers. Although random forest achieved superior AUC, ensemble ANNs obtained comparable AUC with the highest accuracy of all considered models on test data. We analysed possible customer churn causes for a particular financial dataset created at superannuation fund(s). Causal analysis results confirmed variables representing recent SG contribution, annual report and statement preference changed, account growth rate, and balance amount were identified as confounding factors for customer churn with a high degree of belief. The churn rate can be reduced by $\sim 3\%$ for customers with active account > 1year, consistent with expert knowledge. Furthermore, the probability of churn is decreased by ~9% when the customer has a high account balance of over \$100k. A natural progression of this work is extending pattern mining techniques with smaller outcome windows that should be investigated to obtain more efficient prediction results in future studies. Furthermore, different methods to identify causes of churn based on counterfactual causal analysis should be investigated.

# Abbreviations:

| Abbreviations | Meaning | Page |
|---|---|---|
| ANNs | Artificial Neural Networks | |
| AUC | Area Under the Curve | |
| CCP | Customer Churn Prediction | |
| CHAMP | Churn analysis, modelling, and prediction | |
| CNNs | Convolutional Neural Networks | |
| CRM | Customer Relationship Management | |
| DAG | specific magnetisation | |
| DFF | Deep Learning Feed-Forward | |
| DL | Deep Learning | |
| DSR | Data Subset Refuter | |
| LOLIMOT | Locally Linear Model Tree | |
| MCC | Matthews Correlation Coefficient | |
| ML | Machine Learning | |
| NNs | Neural Networks | |
| PDP | Partial Dependence Plot | |
| PLS | Partial Least Squares | |
| RF | Random Forest | |
| RFE | Recursive Feature Elimination | |
| SGD | Stochastic Gradient Descent | |
| SHAP | SHapley Additive Explanation | |
| SMOTE | Synthetic Minority Oversampling Technique | |

## Acknowledgements
This work is partially supported by the Australian Research Council under grant numbers: DP22010371, LE220100078, DP200101374, and LP170100891.

## Author's Contributions
David Hason Rudd made substantial contributions to conceptualisation, investigation, and methodology, analysis and interpretation of data. Guandong Xu and Huan Huo helped in the revision and gave final approval of the version to be published. Project administration, Guandong Xu and Huan Huo.

## Funding
Not applicable.

## Competing Interests
The authors declare that this research work has non-financial Academic and intellectual competing interests.

## Consent for publication
We hereby grant and assign all rights to Human-Centric Intelligence Systems for publication.

## Ethics approval and consent to participate

The dataset used in this study has been unidentified by the company and has no data ethical or privacy issues.

## Availability of data and materials

The python implementation for the proposed framework and causal analysis result, and visualisation in detail, is available on GitHub (https://github.com/DavidHason/Causal Analysis), to simplify reproducing and improving this study experiment results.

## Authors Information:

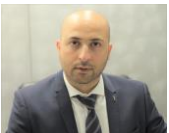

**David Hason Rudd** is currently pursuing a PhD degree at the University of Technology Sydney. He also collaborated with the Data Science Machine Intelligence (DSMI) institute and did several industrial research projects during his PhD study. He has published a few papers in high-rank con- ferences. His main research interests cover churn analysis, prescriptive analytics, data science for business, and machine learning in signal processing.

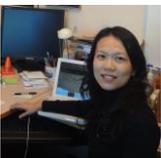

**Huan Huo** is received the B.Eng and Ph.D. degrees from Northeastern University, China in 2002 and 2007, both in Computer Science and Technology. From 2012 to 2014, Angela HUO taught at the Department of Computer Information System, the University of the Fraser Valley in Canada, and did collaborative research in the University of Waterloo as a visiting scholar for one year. Since 2018, she has been a senior lecture in the school of software at the University of Technology Sydney, Australia. Her research interests include data stream management technology, advanced data analysis, and data-driven cybersecurity.

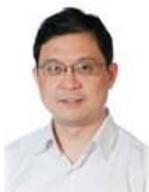

**Guandong Xu** is a Full Professor in Data Science at School of Computer Science and Advanced Analytics Institute, University of Technology Sydney with PhD degree in Computer Science. His research interests cover Data Science, Data Analytics, Recommender Systems, Web Mining, User Modelling, NLP, Social Network Analysis, and Social Media Mining. He has published three monographs in Springer and CRC press, and 220+ journal and conference pa- pers including TOIS, TIST, TKDE, TNNLS, TCBY, TMM, TSE, TSC, TIFS, PR, IJCAI, AAAI, SIGKDD, SIGIR, CVPR, WWW, WSDM, ICDM, ICDE, and CIKM conferences. He is the assistant Editor-in-Chief of World Wide Web Journal and has been serving in editorial board or as guest editors for several international journals, such as TII, TCSS, PR, JBHI, SNAM, JSS, WWWJ, MTAA and OIR. He has received several Awards from academic and industry community. He is elevated the Fellow of Australian Computer Society in 2021.